\documentclass{article}

\usepackage{arxiv}

\usepackage[utf8]{inputenc} 
\usepackage[T1]{fontenc}    
\usepackage{hyperref}       
\usepackage{url}            
\usepackage{booktabs}       
\usepackage{amsfonts}       
\usepackage{microtype}      
\usepackage{amssymb}
\usepackage{amsmath}
\usepackage{xcolor}
\usepackage{cleveref}
\usepackage{subcaption}
\usepackage{graphicx}
\title{A STRATEGY FOR ADAPTIVE SAMPLING OF MULTI-FIDELITY GAUSSIAN PROCESS TO REDUCE PREDICTIVE UNCERTAINTY}

\author{
  Sayan Ghosh\thanks{Address all correspondence to this author, email: \texttt{sayan.ghosh1@ge.com}}, Jesper Kristensen, Yiming Zhang, Waad Subber, Liping Wang\\
  Probabilistics and Optimization Group\\
  General Electric Research \\
  Niskayuna, New York 12309\\
   \\
}

\begin{document}
\maketitle

\begin{abstract}
Multi-fidelity Gaussian process  is a common approach to address the extensive computationally demanding algorithms such as optimization, calibration and uncertainty quantification. Adaptive sampling for multi-fidelity Gaussian process is a changing task due to the fact that  not only we seek to estimate the next sampling location of the design variable, but also the level of the simulator fidelity. This issue is often addressed by including the cost of the simulator as an another factor in the searching criterion in conjunction with  the uncertainty reduction metric. In this work, we extent the traditional design of experiment framework for the multi-fidelity Gaussian process by partitioning  the prediction uncertainty based on the fidelity level and the associated cost of execution. In addition, we utilize  the concept of Believer which quantifies the effect of adding an exploratory design point on the Gaussian process uncertainty prediction. We demonstrated our framework using academic examples as well as a  industrial application of steady-state thermodynamic operation point of a fluidized bed process
\end{abstract}

\section*{INTRODUCTION}
Effective design and optimization of mechanical systems usually require extensive simulation runs and costly physical experiments. As a cost effective alternative, surrogate models have been introduced to approximate the response of the mechanical systems~\cite{shan2010survey, queipo2005surrogate,viana2014special}. Numerical simulations and/or physical experiments are first performed at a given design variables and collected as samples. Surrogate models are then developed based on the scattered samples and serve as the basis for further optimization and uncertainty quantification. With development in the past decades, surrogate models have proved to be a major scheme for effective design optimization and uncertainty quantification of various mechanical systems including but not limited to composite laminates \cite{ zhang2016approaches}, thermodynamic modeling \cite{ honarmandi2019bayesian}, chemo-thermal modeling of composites~\cite{subber2019}, computational fluid dynamics \cite{ bui2008model}, high-performance computing \cite{ zhang2017multi},  structural prognosis \cite{ an2015practical}, crashworthiness-based lightweight design \cite{ zhang2013crashworthiness}, time-dependent reliability design optimization \cite{ hu2016single}, flapping wing design \cite{ chaudhuri2015experimental}. The accuracy of the surrogate model relies on the sampling scheme which systematically determines the location and number of samples.  The sampling scheme could be performed all-at-once or adaptively \cite{haftka2016parallel}. A challenge for sampling scheme is the mixture of dataset with varying fidelity. Fidelity refers to the degree to which the simulations reproduces the response of physical tests. Models with different fidelity could be finite element simulations with changing mesh density, computer simulations with simplified mathematical governing equations \cite{ zhang2017multi} or simulation versus experiments \cite{ zhang2016approaches}. The cost for sampling (i.e. data acquisition) increases with model fidelity. Allocating samples between multi-fidelity models under a fixed budget is expected to enhance the prediction accuracy by making most use of existing information.

Multi-fidelity surrogate models are based on the idea that the high-fidelity experiments can be approximated as a tuned or corrected functions of low-fidelity models \cite{Toropov2001, keane_book,haftka1991,Forrester2007,ghosh2018bayesian,ghosh2015multi}.
A commonly and well known approach to fuse multi-fidelity dataset is by adding a correction term (discrepancy function) on the low-fidelity (LF) dataset towards the high-fidelity (HF) dataset \cite{ KennedyOHagan, zhang2018multifidelity}. The form of discrepancy function assumes the difference between LF and HF models is easier to approximate than the HF model. More details for a comprehensive discussion on MFS could be found in \cite{ park2017remarks, fernandez2016review}. 
The emerging schemes for MFS sampling could be also produced all-at-once or adaptively as the scheme for single-fidelity. All-at-once sampling usually generates first the low-fidelity samples and determines the HF samples as a subset of LF samples \cite{ Gratiet2013}. Reference~\cite{ haaland2010approach} proposed a nested design scheme for categorical and mixed factors. Reference~\cite{ zheng2015difference} performs a comparison study between nested and non-nested design scheme for the effect on MFS accuracy. 
In the context of adaptive sampling, a few initial samples are first produced based on one-time sampling. Then MFS is developed with uncertainty estimation. New samples are recommended from the MFS based on certain infilling criterion to balance between accuracy and cost. Therefore, sampling proceeds alternately between multi-fidelity model. Reference~\cite{ huang2006sequential} proposes adaptive optimization using multi-fidelity kriging. The expected improvement was modified with multiplicative terms to account for fidelity, cost ratio and noise effect. Reference~\cite{ peherstorfer2016multifidelity} adopts LF model to fit the bias distribution for importance sampling. HF model are used to infer the unbiased estimation. Reference~\cite{ chaudhuri2017multifidelity} deals with coupled multidisciplinary system using LF model to approximate the coupling variables and HF model to refine surrogate. 

We found that the adaptive sampling schemes for the multi-fidelity GP surrogate are two step process, where in the first step location or design point on input parameters are determined by finding the location of maximum predictive uncertainty. 
Then in the second step, decision is made whether to run high-fidelity or low-fidelity analysis based on the cost ratio of these analyses \cite{pellegrini2016multi}.
In this work a new criteria is proposed where the selection of next design point as well as fidelity is done in a single step.
In  multi-fidelity GP, the uncertainty on response is coming from both low-fidelity model as well as discrepancy model.
Adding a low-fidelity data reduces uncertainty on low fidelity model and adding a high fidelity  model improves the uncertainty on discrepancy model. 
Therefore hypothesis is that, rather than selecting the next analysis point at maximum overall predictive uncertainty (low-fidelity uncertainty + discrepancy uncertainty), selecting a design point and fidelity where reduction in uncertainty per unit cost is maximum will yield a efficient solution. 
The proposed criteria is also expanded with GP "believer"  \cite{ginsbourger2010kriging}.
The GP believer strategy is based on quantifying the effect of adding an exploratory design point on the multi-fidelity GP uncertainty prediction. For the numerical demonstrations, we consider two illustrative example with different complexity and input dimensions. To illustrate the usefulness of our proposed adaptive schemes, we consider a real industrial application of a steady-state thermodynamic modeling of a fluidized bed process.

\section*{Gaussian Process Surrogate models}

\label{GPSM} 
For many industrial applications, optimization for some operational conditions, model calibration and uncertainty quantification  may require many calls to computationally expansive simulation codes. The computational burden can be overcome by utilizing surrogate models. The  surrogate models require a limited carefully designed simulations through the Design of Experiments (DoE) techniques. Gaussian Process (GP) surrogate model is a common approach  for metamodeling of a wide range of industrial problems~\cite{KennedyOHagan,rasmussen2006,Arendt2012}. The estimation of the prediction uncertainty  in GP  can be considered as its  major desirable property~\cite{KennedyOHagan,rasmussen2006,Arendt2012}. This section outlines the framework surrounding single and multi-fidelity GP. The uncertainty associated with building the GP is discussed in detail. Including properties of this uncertainty is useful in performing adaptive uncertainty sampling.   In adaptive uncertainty sampling technique a new sampling point is added to the training data set where the surrogate model uncertainty is largest.

\subsection*{Single-Fidelity Gaussian Process}

Consider a GP surrogate model of the form:
\begin{equation}
y(x) \sim  GP(m(x),k(x,x' )),
\end{equation}
\noindent where $m(x)$, assumed to be zero here, is the mean function of the input vector $x$, and $k(x,x’)$ is the covariance function. In this work, the covariance function is assumed to be the squared exponential kernel~\cite{rasmussen2006}:

\begin{equation}
k(x,x')=\sigma^2  \exp\left(-\beta (x-x' )^2 \right) + I \lambda^2,
\label{eq:gp_kernel}
\end{equation}
\noindent where $\beta$ are the (inverse) length scale parameters collected in a vector, one per input dimension, $\sigma^2$ captures the data variance as the amount of data variance captured by the model and $\lambda^2$ quantifies the amount of variance captured by the residuals. 
This GP models an output $y(x)$ given an input vector $x$.
Observe now a set of inputs and outputs and collect these in a training data set of $N$ elements $D=\left\{x_i,y_i \right\}_{i=1}^N$.
One property of the GP is that, on any finite set of samples, such as the training data set collected, it reduces to a multivariate Gaussian distribution. Thus, specifically, the GP fitting process translates to fitting the hyperparameters associated with the matrix:
\begin{equation}
K_{(i,j)} = k(x_i,x_j),
\label{eq:gp_cov}
\end{equation}
\noindent where $x_i$ is the $i^{th}$ training datum.

The hyperparameters of the GP are defined as the vector $\theta =(\sigma,\beta,\lambda)$ and need to be fitted to the training data set $D$.
In this work, priors are placed on the hyperparameters to incorporate the initial belief into the data modeling before seeing the data itself, such as smoothness. 
By combining the priors with the likelihood function, the problem of fitting $\theta$ boils down to sampling from the extrema of the posterior distribution $p(\theta|D)$ since, of course, larger values of $p(\theta|D)$ implies more likely models $\theta$. 
The Markov Chain Monte Carlo (MCMC) method both seeks out the extrema and provides a way to sample from it, even in cases where the normalization constant of the posterior probability distribution is unknown.
Incidentally, note that while the GP does have hyperparameters, it is still considered a non-parametric surrogate model since it does not assume any functional form of the data being modeled, such as, e.g., assuming a polynomial.
The MCMC method produces an array, or a chain, of samples of the hyperparameters.
In short, typically, the first  $20-50\%$ of the samples are discarded in order to "lose the memory" of the starting point. After the burn-in, each sample is considered a valid hyperparameter sample from the posterior distribution $p(\theta|D)$.

To simplify the approach ahead, the chain is now condensed into a "lean form" where the median of the chain over each hyperparameter is used to represent the best hyperparameter.  Limitations arise in this method if the distribution over the hyperparameter is multi-modal or cannot well be captured by the median. In any case, we can now think of having a single sample from $p(\theta|D)$ and thus a single GP which in an absolute-deviation sense best fits the data.

\subsection*{Multi-Fidelity Gaussian Process}
Constructing the GP requires limited runs of the computational model at a designed input set. Nevertheless, computational budget allocation might be limited to only a handful of the afforded expansive runs of a high-fidelity simulation code. On the other hand, access to simplified models (low-fidelity) may provide a useful information that at least can capture the general trend of the high-fidelity model. The multi-fidelity GP surrogate can be trained to bridge the information from  various levels of the models complexity~\cite{AlexanderBook}. 

Specifically, consider  the case where we are aiming at modeling two distinct data sets each of a different level of fidelity. We may have a low-fidelity computer simulation that models a given phenomenon, say, the performance of an engine, and the ability to run the real-world experiment. It might be also the case instead of the field experiment, a high-fidelity model is utilized  to simulate the engine performance.

We follow closely Kennedy O’Hagan’s (KOH) methodology~\cite{KennedyOHagan} where the observed data ( i.e., the high-fidelity data), $y(x)$, is represented as a linear combination of a low-fidelity and model a discrepancy term~\cite{KennedyOHagan}:
\begin{equation}
y(x)=\eta(x,\theta)+\delta(x)+ \epsilon,
\label{eq:koh}
\end{equation}
\noindent
where $\theta$ are calibration parameters, i.e., parameters of the low-fidelity model that may or may not have a physical meaning, that can be tuned in order to better match the observed data $y(x)$. Note that Eq.~\ref{eq:koh} contains of two separate GPs . Namely $\eta(x,\theta)$ is a GP for the simulator data and $\delta(x)$  is a GP for capturing the discrepancy between the simulator and the observed data which is collected at the {color{red}independent variable $x$} locations. In this work, the task of calibrating $\theta$ is not considered so these parameters are not included in the model, but future work could include this as well. The noise term $\epsilon$ in  Eq.~\ref{eq:koh}  is assumed as independent and identically distributed zero-mean with constant finite variance Gaussian random variable.

Importantly, each GP in Eq. \ref{eq:koh} is fitted to its own data set. For example, the low-fidelity model $\eta(x,\theta)$ is fitted to a data set $D_{\eta}=\left\{z_i,w_i \right\}_{i=1}^{N_\eta}$ where $z$ is the independent variable and $w$ is the dependent variable (output from the computer simulator). The discrepancy $\delta(x)$ GP is fitted by using information from both the low and the high-fidelity data $D_y=\left\{x_i,y_i \right\}_{i=1}^{N_y}$. The KOH method is a two-part solution: build a base model of the simulation data and a discrepancy model that maps the simulation model to experimental data.

Generally, $z$ and $x$ are not located at the same points, and most typically will not have the same size, i.e., the simulator (low-fidelity model) is run at different input points than the observed (high-fidelity data), but nothing prevents them from being the same.

The covariance matrix on a finite sample of points from both the simulator and the real-world experiment is now given by the following overall structure (the subscript $mf$ refers to multi-fidelity):

\begin{equation}
K_{mf} = \left[
\begin{matrix}
K_y & 0 & 0 \\
0 & K_u & K_{uw} \\
0 & K_{uw}^T & K_w
\end{matrix}
\right],
\end{equation}
\noindent where each covariance matrix has been labeled with a subscript identifying which dependent variable data is being modelled. $K_y$ is the covariance matrix of the high-fidelity data, $K_w$ is the covariance of the low-fidelity data and the newly introduced variable $u$ refers to the low-fidelity data predicted on the high-fidelity points and thus $K_{uw}$ is the covariance matrix between the low-fidelity data and the low-fidelity model predicting high-fidelity data. This covariance matrix contains multiple hyperparameters via the covariance matrices which in turn are defined from the covariance function in Eq. \ref{eq:gp_kernel}. The parameters are fitted with MCMC in the same way as previously discussed.

\subsection*{ Gaussian Process Uncertainty}

Next, we discuss the uncertainty estimation associated with the GP predictions. The prediction uncertainty is a crucial aspect of GP and sets it apart from many other surrogate modeling techniques. The following observations regarding the GP prediction uncertainty (variance) will be important for the developments to follow:

\begin{enumerate}
	\item The prediction uncertainty increases to the variance of the training data far away from the training data. Thus, there is a lower bound of 0 and an upper bound given by the variance of the data.
	\item The prediction uncertainty is small near training data points. On a training datum the uncertainty is as small as it gets anywhere else.
	\item The prediction uncertainty can be non-zero at a training datum in our formulation, but GPs can be used as interpolators if modifying the kernel in Eq. \ref{eq:gp_kernel}. 
\end{enumerate}

Incidentally, consider another surrogate modeling techniques such as the Neural Network (NN) \cite{bishop2006pattern}. One way to obtain the uncertainty in NNs is by randomly selecting multiple subsets of the data fitting an NN on each and combining all the NNs into a “mean NN” and its associated uncertainty. In this case, there is no guarantee that the properties above hold true, but as we shall see, property 2 is especially helpful. 

To predict the mean and uncertainty of the GP at any point, generally referred to as an unseen point (but is allowed to be in the training set), $x^{*}$, the first step is to compute the covariance vector between the new point and the existing training data $k(x^{*},x)=k^{*}$. Then, the predicted mean $m^{*}$ value and uncertainty $V^{*}$ associated with the GP are then given by:

\begin{align}
m^*&={k^*}^T K^{-1} y, \\
V^*(x^*)&=k(x^*,x^* )-{k^*}^T K^{-1} k^*,
\label{eq:gp_var}
\end{align}
\noindent
where $K$ is given in Eq. \ref{eq:gp_cov} and $k(x^*,x^* )$ is the covariance of the unseen point.

\section*{Sampling Strategies}
Constructing a multi-fidelity GP surrogate requires a careful experimental design of the sampling strategy. The design of the experiment (DoE) should provide an optimal selection of  the  design variables $x \in D$ such that it emulates the behavior of the expansive response of the high-fidelity model while minimizing a computational cost or performance metrics.  Note that the selection criterion is defined by an objective function that can be maximized or minimized depending on the  goal of the experiment. For multi-fidelity GP, the design space $D=\{D_\eta \cup D_y\}$ consists of low and high fidelity input variables, and thus an adaptive sampling strategy is more appropriate to achieve the aim of the experiment. In adaptive sampling, the design space is augmented  sequentially by adding a new set of the  variables that optimally satisfy the objective criterion. In addition for the multi-fidelity framework, the next design set not only consists of the best location to perform the computer experiment, but also the level of fidelity (i.e, execution of low or high fidelity model). The decomposition property of the prediction variance into a contribution from the low fidelity model and the discrepancy term provides a strategy to determine the next level of model fidelity to simulate at the next optimal design point~\cite{Gratiet2013}. In the next sections, we discuss the adaptive sampling strategy for the single-fidelity  GP, and we propose a new adaptive sampling techniques for the multi-fidelity  GP.


\subsection*{Adaptive sampling for the single-fidelity  GP}

In adaptive sampling for the single-fidelity  GP, typically, we start with a space-filling DoE with a number of points depending on the size of the problem~\cite{AlexanderBook}. Then, iteratively, one or multiple points are added to the initial design and the GP is re-fitted in each iteration. For example, say we start with 10 points in the design and add points until our training data has 50 points. The question is which points should be chosen in each iteration? If the task is to learn the posterior distribution of the hyperparameters as well as possible, or in other words, learn the response surface to some threshold degree of accuracy throughout the design space, then a method called uncertainty sampling is an approach to take \cite{liu2009self}.

In each iteration of uncertainty sampling, the next point to be added to the design space is the point with the largest predictive variance of the GP. The predictive variance is given in Eq. \ref{eq:gp_var}, and thus uncertainty sampling picks the next point $x^*$ believed to be the most informative point as:
\begin{equation}
x^* = \operatorname*{arg\,max}_x V^*(x), 
\end{equation}
\noindent
where the right hand side seeks to find the input point $x$ that maximizes a utility function, here simply $V$, but utility functions can be more complex. Other approaches, for example, include variance reduction in which the total volume of the confidence band is sought to be reduced the most, worst-case variance reduction in which the change in maximum GP uncertainty is the utility function among others~\cite{kristensen2016expected}.

Assume that the resources required to obtain a new point is constant across the input space. In this case, the concept of the cost associated with selecting a specific point is not relevant. This will change in the sections to follow as we consider multi-fidelity GPs.


\subsection*{Adaptive Sampling in Multi-fidelity GP}
\begin{figure}
		\centering
		\includegraphics[trim=0 0 500 0, clip,width=4in]{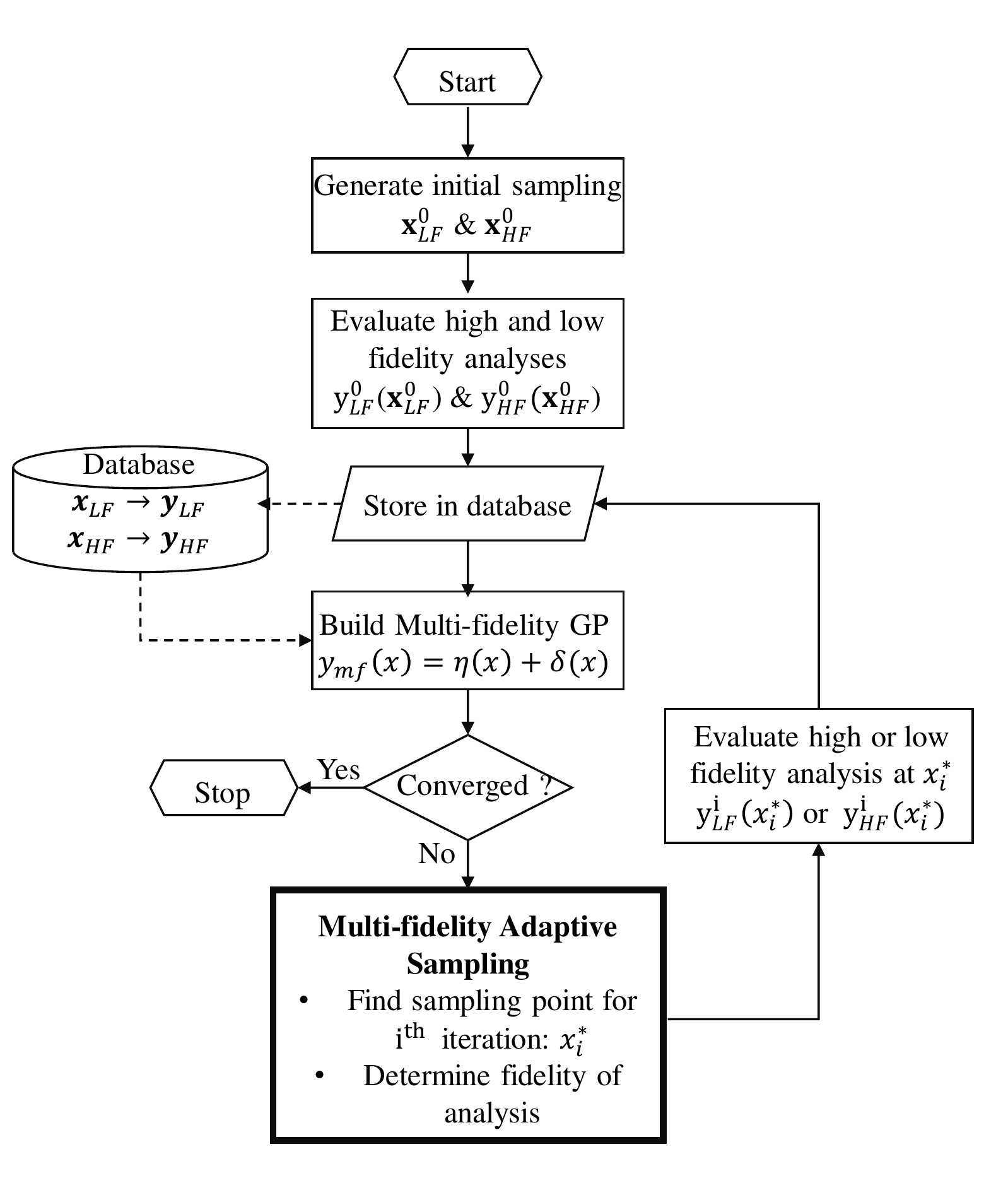}
	\caption{A TYPICAL MULTI-FIDELITY ADAPTIVE SAMPLING PROCESS}
	\label{fig:mfs_process}
\end{figure}

A typical multi-fidelity adaptive sampling process is shown in Fig. \ref{fig:mfs_process}.
The process starts by generating initial samples of input variables for both low and high-fidelity analysis. 
Next, is to carry of low and high-fidelity analysis to evaluate the response and generating the initial database.
The number of initial samples is another factor which can affect the speed of convergence and total cost of the process, however that is not considered to be the part of the current study.
Next, a multi-fidelity model is built by building a model for low-fidelity model ($\eta(x)$) and the discrepancy model ($\delta(x)$) for the discrepancy between low and high-fidelity data.
Then the model convergence is checked with provided validation metric to evaluated if the model is accurate enough as per requirement. 
If not, then multi-fidelity adaptive sampling strategy is applied. 

Multi-fidelity sampling strategy tries to answer two new issues:

\begin{enumerate}
	\item In a given iteration, should we choose to run the low-fidelity model or the high-fidelity model? How do we decide?
	\item How do we factor in the difference in costs of running the models of varying fidelities?
\end{enumerate}

Once the next sampling point, $x^*$, and the fidelity of analysis is determiner, new analysis is carried out at $x^*$ to evaluate $y_{LF}(x^*)$ or $y_{HF}(x^*)$.
The new data is stored in the database and process is repeated by building a new multi-fidelity GP with the updated database.

The focus of the current work is the multi-fidelity adaptive sampling strategy as shown by the bold box in Fig. \ref{fig:mfs_process}.
Typically, a two step strategy is used in  multi-fidelity adaptive sampling strategy \cite{pellegrini2016multi}, where in the first step $x^*$ is determined without considering the cost impact. 
Once $x^*$ is determined, then fidelity of analysis is decided based on the cost ratio between high and low fidelity analyses. 
In this paper we refer this method as  sampling at Maximum Multi-Fidelity Uncertainty to Cost Ratio (Max MF-UCR) and as considered as baseline method to compare the new proposed approach.

The proposed approach, sampling at Maximum Individual-Fidelity Uncertainty to Cost Ratio (Max IF-UCR), carries out one step process where it determines the next sampling point as well the fidelity of analysis in a single step by considering the cost of analyses during the selection process of $x^*$. 
The proposed approach is also extended to sampling at Maximum Individual-Fidelity Uncertainty to Cost Ratio using Believer (Max IF-UCR Bel) to study if any benefit can be achieved by using GP believer.
Details of each strategy is given below:

\subsubsection*{Sampling at Maximum Multi-Fidelity Uncertainty to Cost Ratio (Max MF-UCR)}
In this strategy, the sampling at next iteration is carried out at design $x^*$ where the predictive uncertainty of multi-fidelity response ($\sigma_{y_{mf}})$ is maximum, i.e.

\begin{equation}
x^* = \operatorname*{arg\,max}_x \sigma_{y_{mf}},
\end{equation}
\noindent where $\sigma_{y_{mf}}$ is the standard deviation of response $y_{mf}$ of multi-fidelity GP.
Let's say $C_H$ is the cost of each high-fidelity analysis and $C_L$ is the cost of low fidelity analysis then, if $\frac{\sigma_{\eta}(x^*) }{C_L} \ge \frac{\sigma_{\delta}(x^*) }{C_H}$, then low fidelity analysis is carried out at $x^*$ during the next iteration, otherwise high-fidelity is carried out.

\subsubsection*{Sampling at Maximum Individual-Fidelity Uncertainty to Cost Ratio (Max IF-UCR)}
In this strategy, the sampling at next iteration is carried out at design $x^*$ where the uncertainty reduction per unit cost is maximum. At a given $x$, uncertainty reduction per unit cost for low-fidelity is function of $\eta(x)$, i.e. $\sigma_{\eta}(x)/C_L$. Similarly, for high-fidelity uncertainty reduction per unit cost is function of $\delta(x)$, i.e. $\sigma_{\delta}(x)/C_H$.
The next sampling point is chosen as:

\begin{equation}
x^* = \operatorname*{arg\,max}_x  \left[ \max \left( \frac{\sigma_{\eta}(x) }{C_L}, \frac{\sigma_{\delta}(x) }{C_H}  \right) \right],
\end{equation}
\noindent where $\sigma_{y_{mf}}$ is the standard deviation of response $y_{mf}$ of multi-fidelity GP.
If $\frac{\sigma_{\eta}(x^*) }{C_L} \ge \frac{\sigma_{\delta}(x^*) }{C_L}$, then low fidelity analysis is carried out at $x^*$ during the next iteration, otherwise high-fidelity is carried out.

\subsubsection*{Sampling at Maximum Individual-Fidelity Uncertainty to Cost Ratio using Believer (Max IF-UCR Bel)}
GP believer is the concept of quantifying the impact on the GP model from adding a hypothetical new data point. In other words, consider a multi-fidelity GP which has been built on the joint training data set $D=(D_{\eta},D_{y})$. Consider an unseen point $x$: how much does the overall uncertainty reduce by if running the low-fidelity code at $x$? How much does this compare to running the high-fidelity code?

To gauge the effect of adding a hypothetical point to $D$, we observe that the variance of the multi-fidelity model does not need the observed value $y^*$  of the unseen datum $(x^*, y^*)$. In fact:

\begin{equation}
V_{mf}^* (x^* )=k_{mf} (x^*,x^* )- k{*T}_{mf}K_{mf}^{(-1)} k_{mf}^*.
\end{equation}

One can add $x^*$ to the covariance matrices and evaluate the new variance. However, this ignores the fact that adding a datum technically requires re-fitting of the GP since the hyperparameters depend on the training data. The fitting process is typically not very expensive compared to running the low- or high-fidelity models so can be done. The issue is that the number of unseen points we have to gauge can be on the order of $10-100,000$. If the fitting takes 10 seconds the process of selecting a single new point can take on the order of $1-10$ days. We are looking to spend much less time, on the order of seconds, on this task.

A solution to this can be to assume that the hyperparameters fitted with MCMC are practically speaking unchanged temporarily while searching for the next point. Thus, re-fitting is not required. More advanced methods could be envisioned here leveraging the MCMC chain in other ways.
When gauging the effect of adding a low-fidelity point, only the matrices involving low-fidelity data are of course changed and similarly for a hypothetical high-fidelity point.

Similar to Max IF-UCR, Max IF-UCR-Bel uses GP believer to determine how much is the uncertainty reduction if a low-fidelity or high-fidelity analysis is carried out at given point $x$. 
The next sample and the fidelity is then chosen for which the uncertainty reduction per unit cost is maximum as: 
\begin{equation}
\begin{aligned}
x^* &= \operatorname*{arg\,max}_x   \left[ \max \left( \frac{\sigma_{y}(x) - \sigma_y^{Bel}(x | x^{LF}_{bel} =x)}{C_L}, \right.\right. \\ & \left. \left. \frac{\sigma_{y}(x) - \sigma_y^{Bel}(x | x^{HF}_{bel} =x)}{C_H}  \right) \right],
\end{aligned}
\end{equation}

\noindent where $\sigma_y^{Bel}(x | x^{LF}_{bel} =x)$ is the standard deviation of multi-fidelity predictor $y$ at $x$ when a low-fidelity "believer" is added at  $x^{LF}_{bel} =x$. Similarly, $\sigma_y^{Bel}(x | x^{HF}_{bel} =x)$ is  the standard deviation of multi-fidelity predictor $y$ at $x$ when a high-fidelity "believer" is added at  $x^{HF}_{bel} =x$.
In the next iteration, sampling is done using low fidelity analysis if $ \frac{\sigma_{y}(x^*) - \sigma_y^{Bel}(x^* | x^{LF}_{bel} =x^*)}{C_L} \ge \frac{\sigma_{y}(x^*) - \sigma_y^{Bel}(x^* | x^{HF}_{bel} =x^*)}{C_H}$. else high-fidelity sampling is carried out at $x^*$.
\subsection*{Test Problem}

\subsubsection*{1-D Forrester Function}
The first test problem consists of analytical functions \cite{forrester2008engineering} to define both high and low fidelity analyses. 
The high fidelity equation is given as
\begin{equation}
f_H(x) = (6x - 2)^2 \sin(12x-4).
\end{equation}
The function is one-dimensional and is multi-modal in nature and has been used in literature for validating and testing surrogate models for single and multi-fidelity. 

The low fidelity equation is given as
\begin{equation}
f_L(x)  = Af_H(x) + B(x-0.5) - C, 
\end{equation}
\noindent where $A=0.6$, $B = 10$ and $C = 7$ has been used in the current study. 
Both the high and low fidelity functions are evaluated for $x\in [0,1]$.

To start the multi-fidelity adaptive sampling, $4$ samples for low-fidelity and $2$  samples for high-fidelity are randomly generated in domain on $x\in [0,1]$ and evaluated using the respective functions.
Additionally, two "prospective" databases are generated for each of the high-fidelity and low-fidelity analyses. 
These databases contains prospective $100$ random samples of input $x$, which will be used to evaluate adaptive sampling criteria to pick the next sampling point and fidelity at each iteration. 
It should be noted that the design point in each of these databases are not collocated.
Also, additional $100$ samples are generated and evaluated using high-fidelity analysis to estimate error statistics. 

Multi-fidelity adaptive sampling process is started by building multi-fidelity GP using the initial samples.
At the end of each iteration. new sample $x^*$ either from low-fidelity or high-fidelity is picked from the "prospective" databases based on the aforementioned strategy.
Based on the determined fidelity, analyses is carried out at $x^*$ and is added to GP database.
The multi-fidelity GP is then retrained with updated database. The process is carried out for $10$ sampling iterations. 

\begin{figure}[h]
	\begin{center}
		\begin{subfigure}[t]{0.45\textwidth}
			\includegraphics[width=\textwidth]{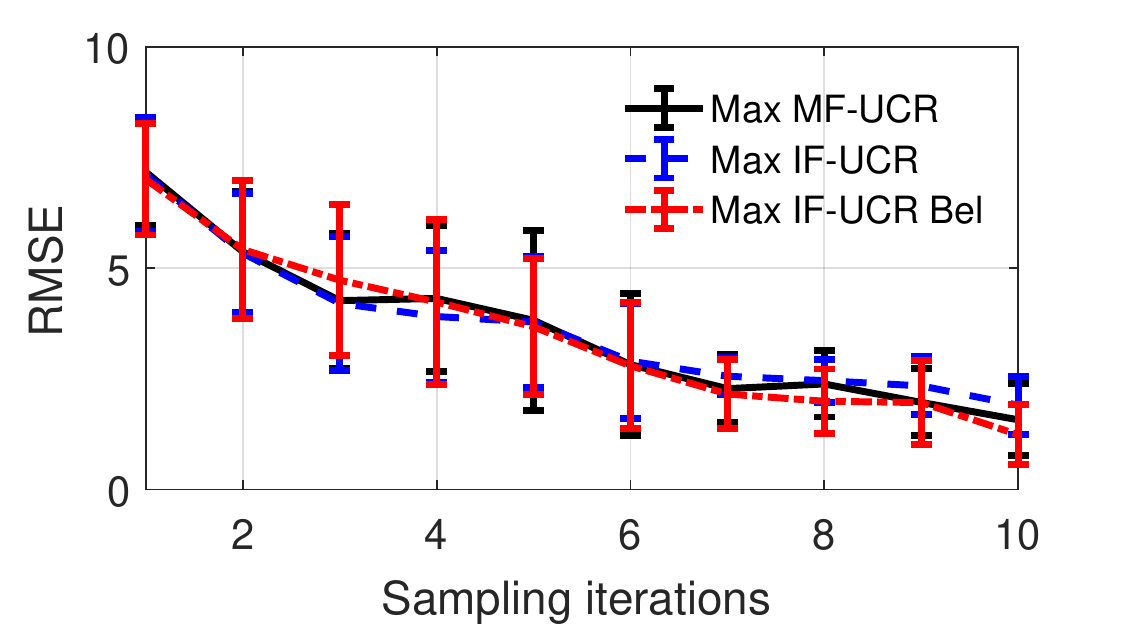}
			\caption{ROOT MEAN SQUARED ERROR (RMSE) AS FUNCTION OF SAMPLING ITERATION}
			\label{fig:1D_1_2a}
		\end{subfigure}
		\begin{subfigure}[t]{0.45\textwidth}
			\includegraphics[width=\textwidth]{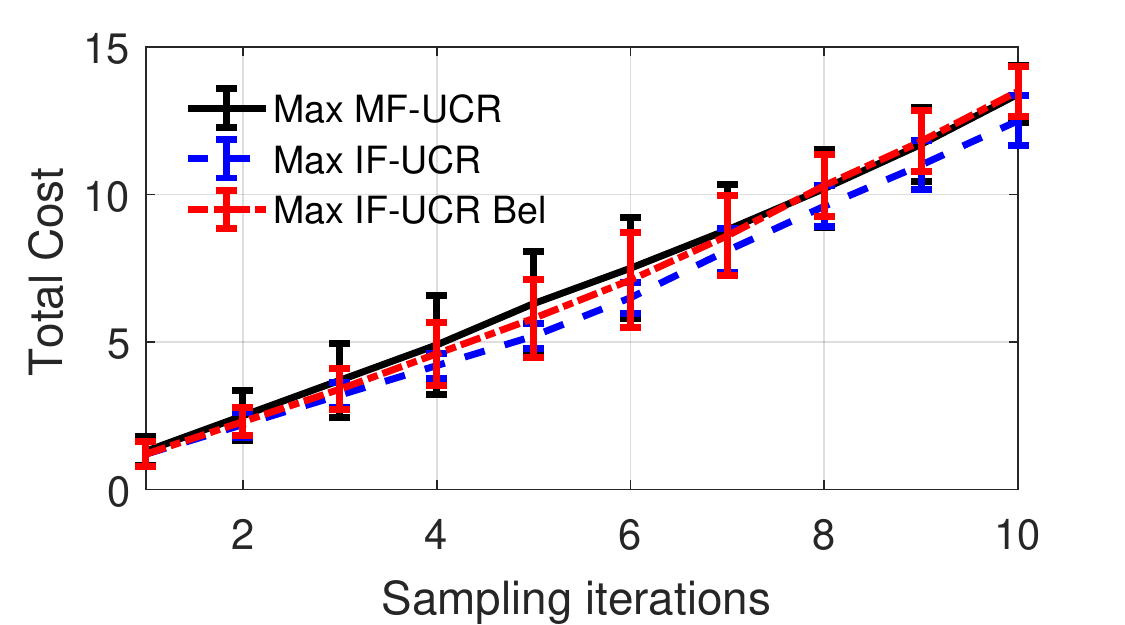}
			\caption{TOTAL COST AS FUNCTION OF SAMPLING ITERATION}
			\label{fig:1D_1_2b}
		\end{subfigure}
	\end{center}
	\caption{COMPARISON OF METHODS FOR 1-D FORRESTER FUNCTION WITH COST RATIO OF HIGH-FIDELITY AND LOW-FIDELITY OF 2:1}
	\label{fig:1D_1_2}
\end{figure}

\begin{figure}[h]
	\begin{center}
		\begin{subfigure}[t]{0.45\textwidth}
			\includegraphics[width=\textwidth]{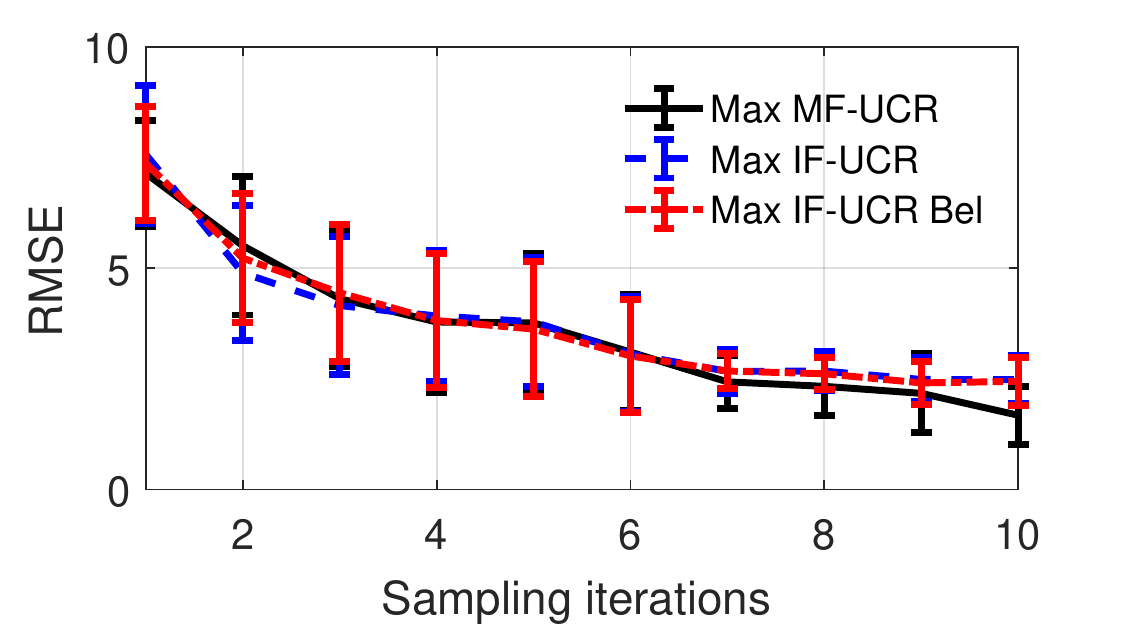}
			\caption{ROOT MEAN SQUARED ERROR (RMSE) AS FUNCTION OF SAMPLING ITERATION}
			\label{fig:1D_1_5a}
		\end{subfigure}
		\begin{subfigure}[t]{0.45\textwidth}
			\includegraphics[width=\textwidth]{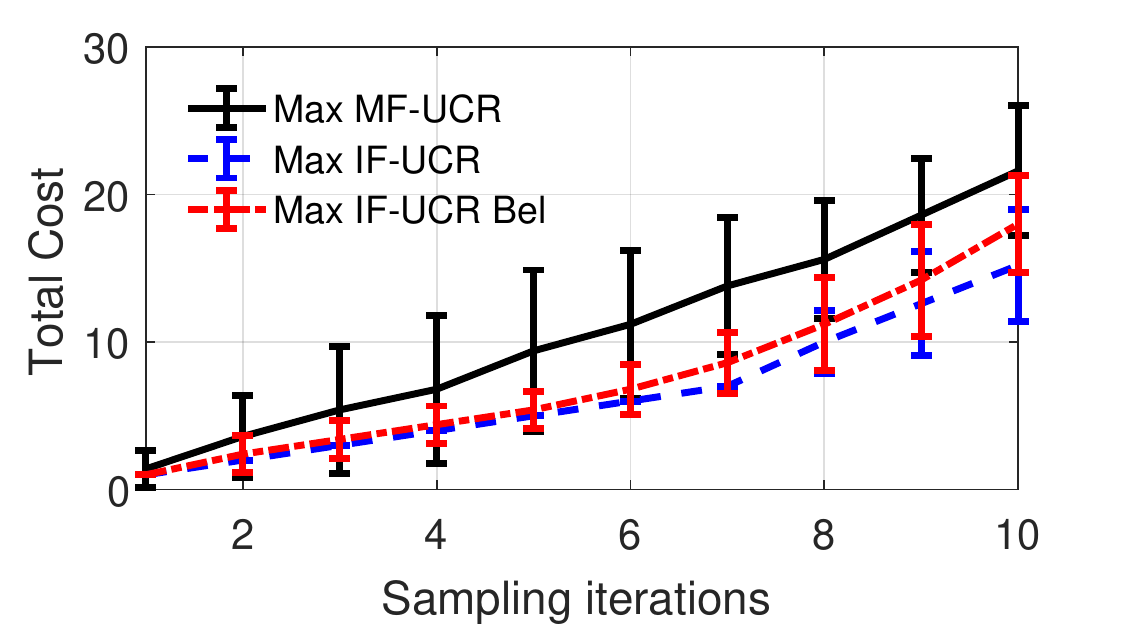}
			\caption{TOTAL COST AS FUNCTION OF SAMPLING ITERATION}
			\label{fig:1D_1_5b}
		\end{subfigure}
	\end{center}
	\caption{COMPARISON OF METHODS FOR 1-D FORRESTER FUNCTION WITH COST RATIO OF HIGH-FIDELITY AND LOW-FIDELITY OF 5:1}
	\label{fig:1D_1_5}
\end{figure}

\begin{figure}[h]
	\begin{center}
		\begin{subfigure}[t]{0.45\textwidth}
			\includegraphics[width=\textwidth]{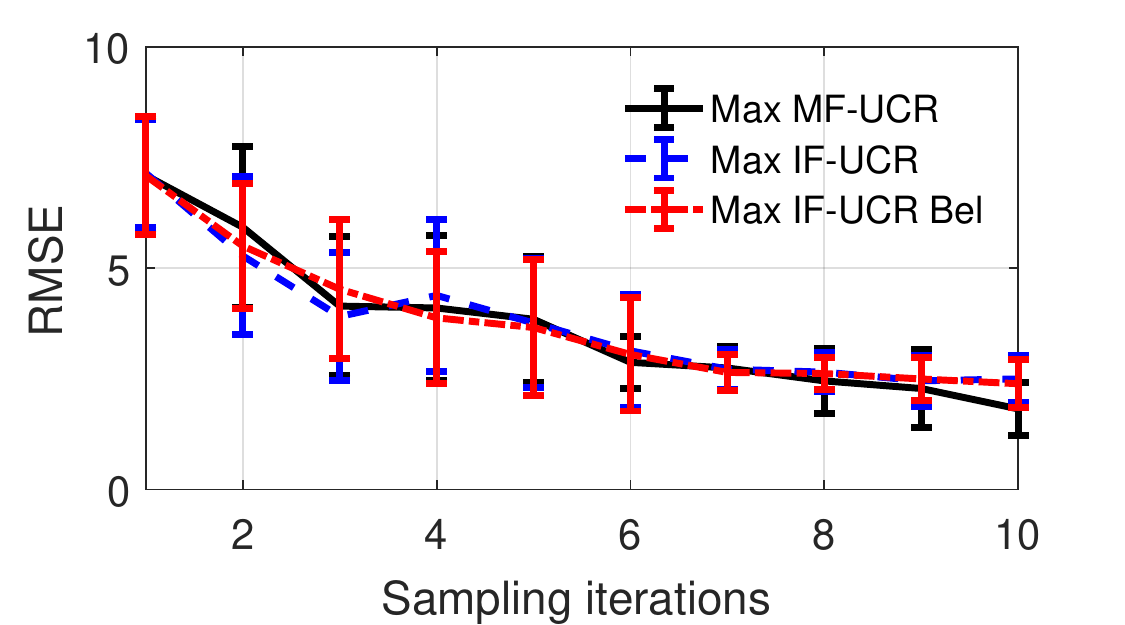}
			\caption{ROOT MEAN SQUARED ERROR (RMSE) AS FUNCTION OF SAMPLING ITERATION}
			\label{fig:1D_1_10a}
		\end{subfigure}
		\begin{subfigure}[t]{0.45\textwidth}
			\includegraphics[width=\textwidth]{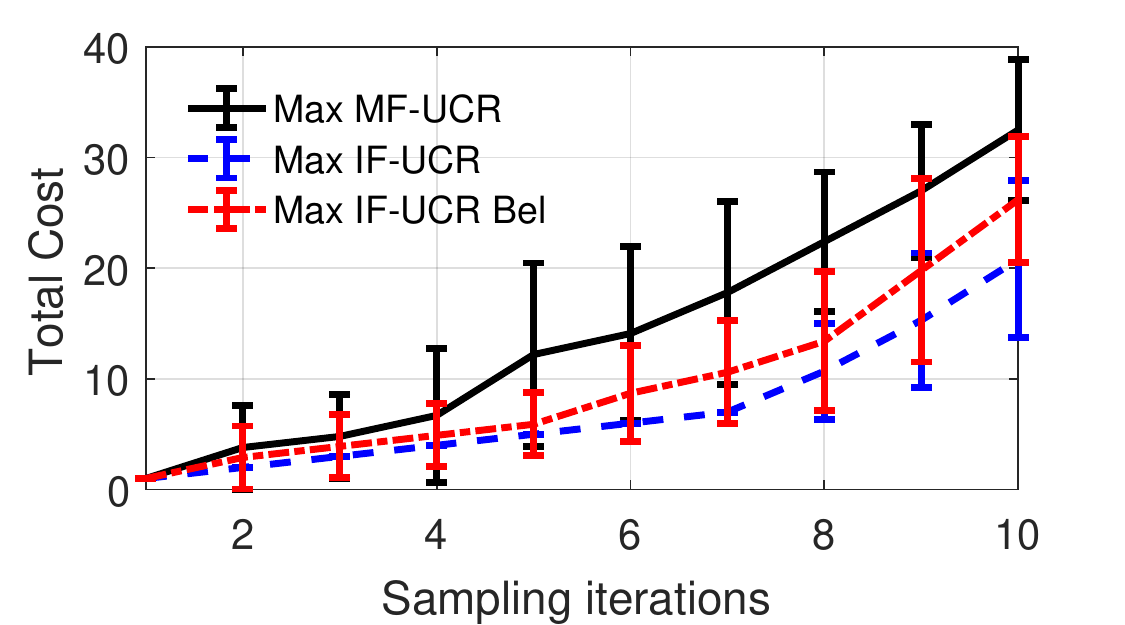}
			\caption{TOTAL COST AS FUNCTION OF SAMPLING ITERATION}
			\label{fig:1D_1_10b}
		\end{subfigure}
	\end{center}
	\caption{COMPARISON OF METHODS FOR 1-D FORRESTER FUNCTION WITH COST RATIO OF HIGH-FIDELITY AND LOW-FIDELITY OF 10:1}
	\label{fig:1D_1_10}
\end{figure}

The overall process is repeated $10$ times, where in each case different initial sample is chosen to carry out the multi-fidelity adaptive sampling.
This is done to verify that overall selection criteria is robust to different initial samples. 
Also, three scenarios are studied with different high-fidelity to low-fidelity cost ratio ($C_H:C_L$) of $2:1, 5:1$, and $10:1$.
The results of these scenarios are shown in Fig. \ref{fig:1D_1_2}, \ref{fig:1D_1_5} and \ref{fig:1D_1_10}.
In each of these figures, the first plot shows the convergence of root means squared error (RMSE) with respect to adaptive sampling iterations, with error bars showing the uncertainty across $10$ different runs. 
The second subplot shows the total cost of analysis with respect to sampling iteration.
As observed in all the cases the RMSE converges (Fig. \ref{fig:1D_1_2a}, \ref{fig:1D_1_5a} and \ref{fig:1D_1_10a}) to similar value after $10$ iteration for each of the multi-fidelity selection criteria.
For the scenario of $C_H:C_L = 2:1$ (Fig. \ref{fig:1D_1_2b}), although we found the total cost at the end of $10$ iteration of was not significant different.
However, for the scenario of $C_H:C_L = 5:1$ and $10:1$ (Fig. \ref{fig:1D_1_5b} and \ref{fig:1D_1_10b}), both Max IF-UCR and Max IF-UCR Bel criteria did better than Max MF-UCR. 
It was also observed that Max IF-UCR to be better than Max IF-UCR Bel criteria for both these scenarios. 
We also found that the as the $C_H:C_L$ increases, the cost reduction for both  Max IF-UCR and Max IF-UCR increases when compared to Max MF-UCR.

\subsubsection*{ 4-D Park Function}

The second numerical experiment is carried out with $4$-dimensional analytical test functions.
The high fidelity function is given in Eq. \ref{eq:park_hf} and was  used by Park and Cox \cite{park1991tuning,cox2001statistical} for testing method for tuning computer code.

\begin{equation}
f_H(x)  = \frac{x_1}{2} \left[\sqrt{1 + (x_2 +x_3^2)\frac{x_4}{x_1^2}} - 1 \right] + (x_1 + 3x_4)\exp\left[1 + \sin(x_3)\right]
\label{eq:park_hf}
\end{equation}

The low-fidelity analysis is represented by Eq. \ref{eq:park_lf} and was used by Xiong et al. \cite{xiong2013sequential}. 
The input domain for both low and high-fidelity analysis is $x \in (0,1)$.

\begin{equation}
f_L(x) = \left[1 + \frac{\sin(x_1)}{10}\right] f_H(x) - 2x_1 + x_2^2 + x_3^2 + 0.5
\label{eq:park_lf}
\end{equation}

At first, two "prosective" databases are generated for each of the high-fidelity and low-fidelity analyses. 
These database contains $100$ random samples of input $x$, which will be used to evaluate adaptive sampling criteria to pick the design point and fidelity at each iteration. 
Also, additional $100$ samples are generated and evaluated using high-fidelity analysis to estimate error statistics.
The multi-fidelity experiments begins with using $2$ samples from high-fidelity analyses and $4$ samples from low-fidelity analysis, randomly selected in $x \in (0,1)$, and building an initial mult-fidelity GP on these. 

\begin{figure}h]
	\begin{center}
		\begin{subfigure}[t]{0.45\textwidth}
			\includegraphics[width=\textwidth]{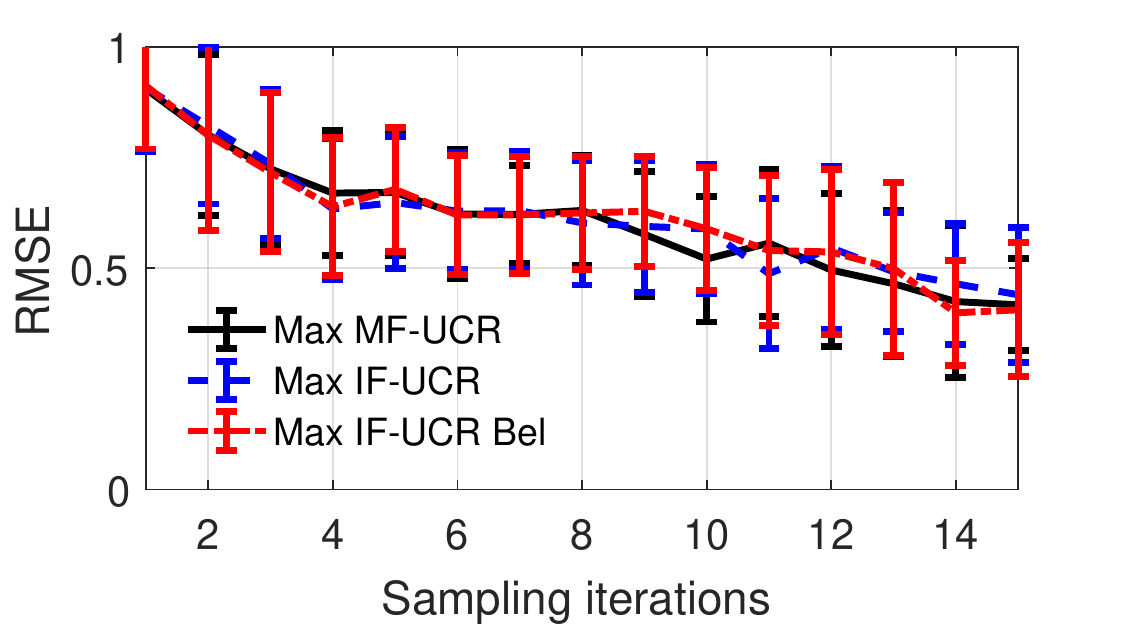}
			\caption{ROOT MEAN SQUARED ERROR (RMSE) AS FUNCTION OF SAMPLING ITERATION}
		\end{subfigure}
		\begin{subfigure}[t]{0.45\textwidth}
			\includegraphics[width=\textwidth]{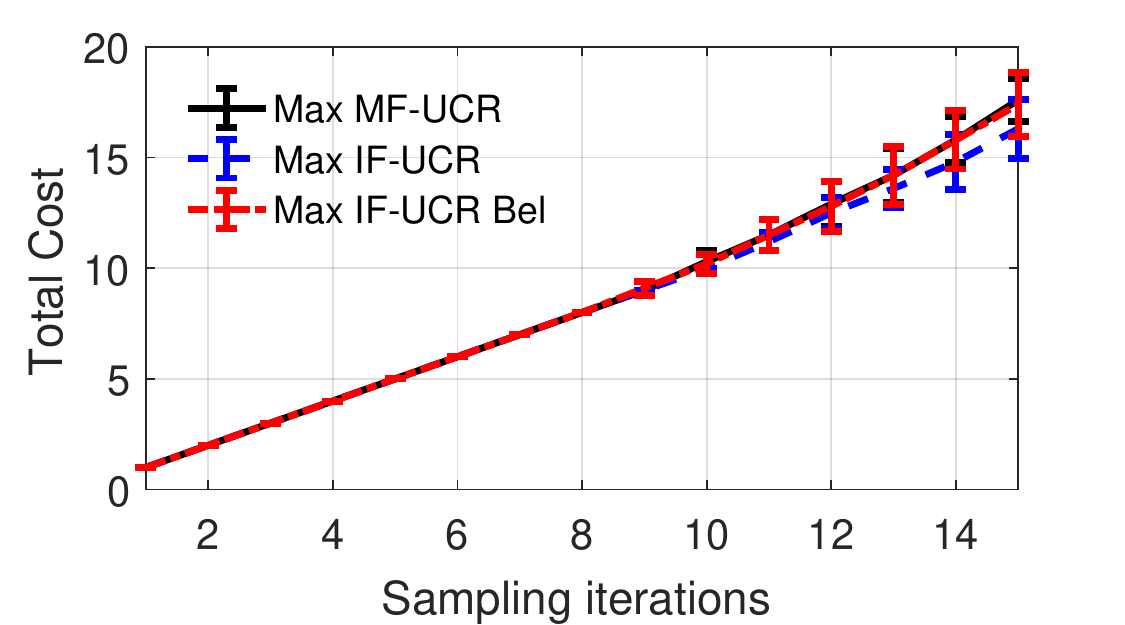}
			\caption{TOTAL COST AS FUNCTION OF SAMPLING ITERATION}
		\end{subfigure}
	\end{center}
	\caption{COMPARISON OF METHODS FOR 4-D PARK FUNCTION WITH COST RATIO OF HIGH-FIDELITY AND LOW-FIDELITY OF 2:1}
	\label{fig:4D_1_2}
\end{figure}

\begin{figure}[h]
	\begin{center}
		\begin{subfigure}[t]{0.45\textwidth}
			\includegraphics[width=\textwidth]{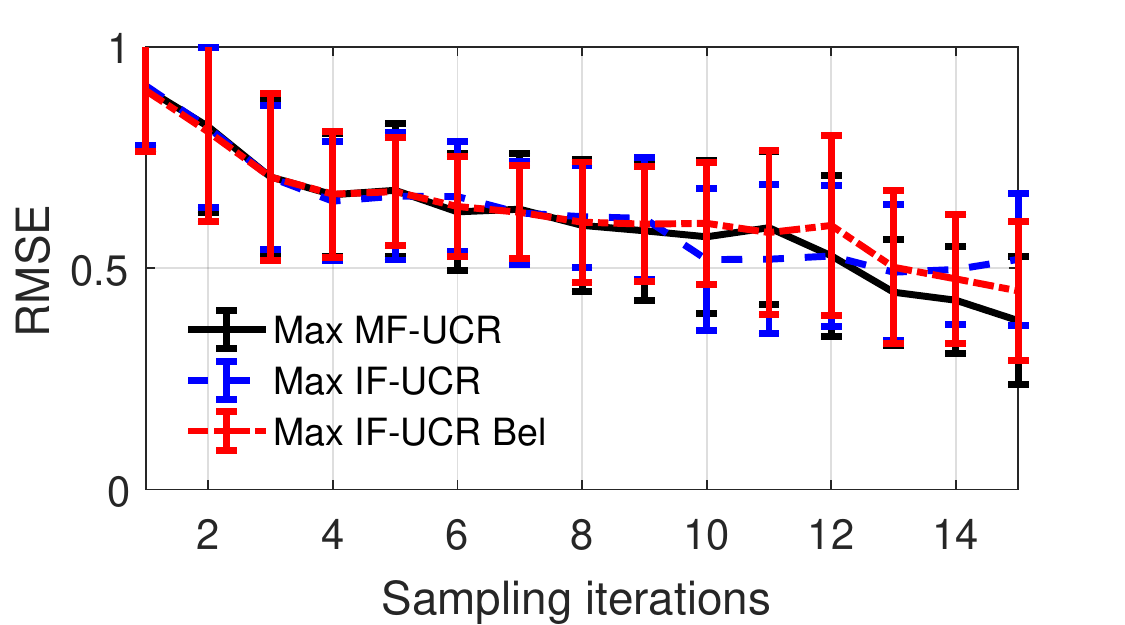}
			\caption{ROOT MEAN SQUARED ERROR (RMSE) AS FUNCTION OF SAMPLING ITERATION}
		\end{subfigure}
		\begin{subfigure}[t]{0.45\textwidth}
			\includegraphics[width=\textwidth]{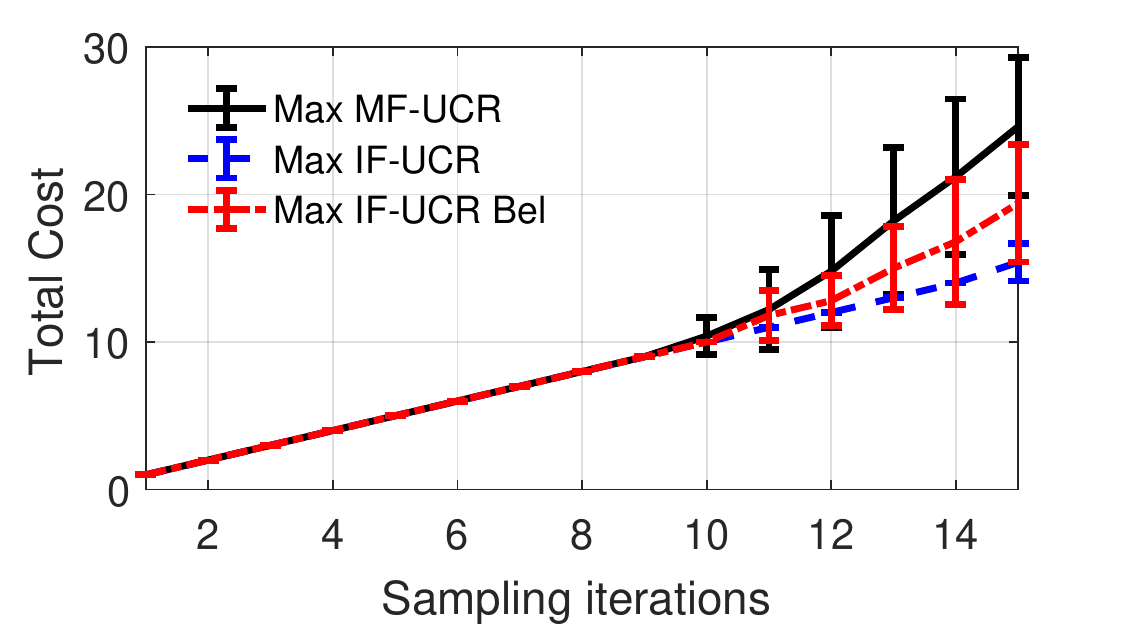}
			\caption{TOTAL COST AS FUNCTION OF SAMPLING ITERATION}
		\end{subfigure}
	\end{center}
	\caption{COMPARISON OF METHODS FOR 4-D PARK FUNCTION WITH COST RATIO OF HIGH-FIDELITY AND LOW-FIDELITY OF 5:1}
	\label{fig:4D_1_5}
\end{figure}

\begin{figure}[h]
	\begin{center}
		\begin{subfigure}[t]{0.45\textwidth}
			\includegraphics[width=\textwidth]{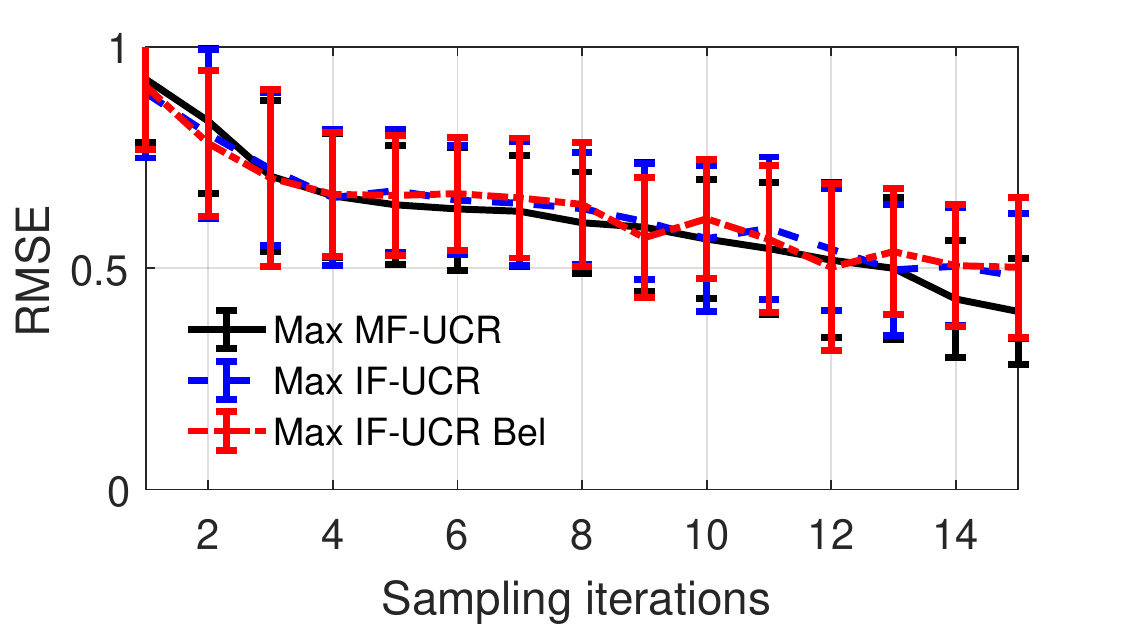}
			\caption{ROOT MEAN SQUARED ERROR (RMSE) AS FUNCTION OF SAMPLING ITERATION}
		\end{subfigure}
		\begin{subfigure}[t]{0.45\textwidth}
			\includegraphics[width=\textwidth]{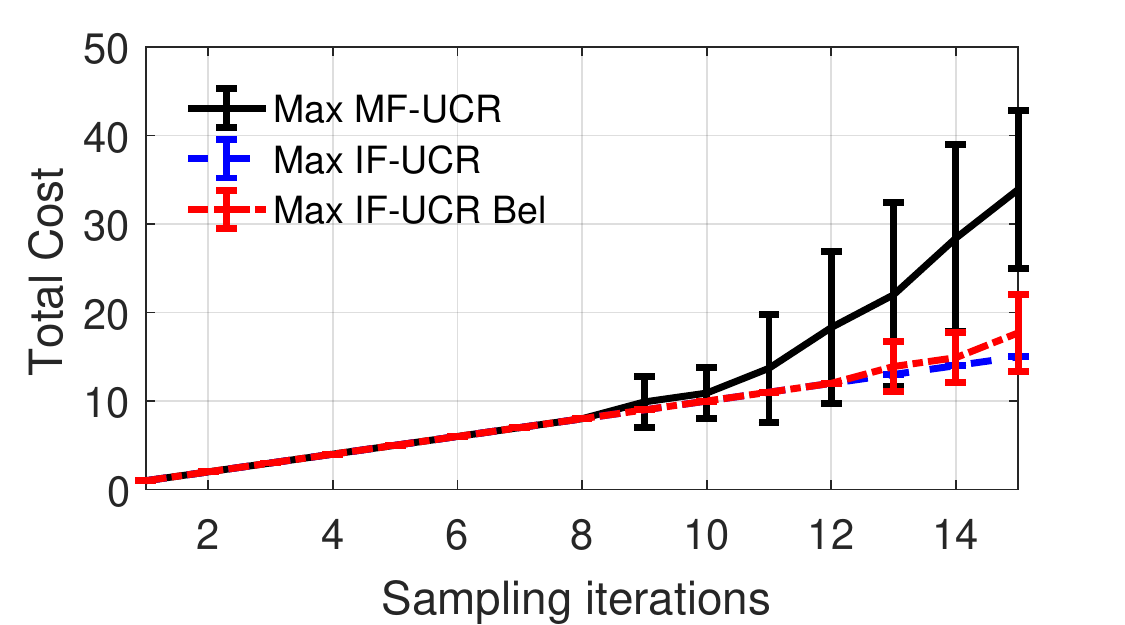}
			\caption{TOTAL COST AS FUNCTION OF SAMPLING ITERATION}
		\end{subfigure}
	\end{center}
	\caption{COMPARISON OF METHODS FOR 4-D PARK FUNCTION WITH COST RATIO OF HIGH-FIDELITY AND LOW-FIDELITY OF 10:1}
	\label{fig:4D_1_10}
\end{figure}

Similar to $1-D$ test problem, three scenarios are studied with different high-fidelity to low-fidelity cost ratio ($C_H:C_L$) of $2:1, 5:1$, and $10:1$, and for each scenario $10$ cases were carried out with different initial sampling. 
The results for each scenario are shown in Fig. \ref{fig:4D_1_2}, \ref{fig:4D_1_5}, and \ref{fig:4D_1_10}.

In all the scenarios, RMSE converged to similar value after $15$ iterations for all the adaptive sampling approaches. 
In terms of cost, total cost was significantly different for any of the methods for $C_H: C_L = 2:1$. 
At $C_H: C_L = 5:1$, Max IF-UCR performed the best, followed by Max IF-UCR Bel.
At $C_H: C_L = 10:1$, both Max IF-UCR and Max IF-UCR Bel performed similarly and much better than Max MF-UCR.

\subsubsection*{ 6-D Fluidized-Bed Problem}
For this test, dataset was taken from the study carried out by Dewettinck et al. \cite{dewettinck1999modeling} on thermodynamic modeling of top-spray fluidized bed, which was used to understand the impact of process variables and ambient changes known as the so-called weather effect.
The quantity of interest is the temperature of steady-state operation of fluidized-bed, which is function of six variables:  humidity ($H_R$), room temperature ($T_R$), temperature of the air from the pump ($T_a$), flow rate of the coating solution ($R_f$), pressure of atomized air ($P_a$), and fluid velocity of the fluidization air ($V_f$).
In that study, three different fidelity of the model has been used and the results were validated with experimental results for $28$ different operating conditions. 
In the present work, $8$ out $28$ data points were kept aside to carry out validation and estimating the error statistics. 
From the remaining data, $2$ data were randomly chosen from experimental data as high-fidelity analysis and $4$ data from mid-fidelity data were used as low-fidelity data as starting point (first iteration) of adaptive sampling process. 
During the adaptive sampling process, the new designed point were selected from remaining data in the database.
Ten cases were runs to analyze the robustness of the method with respect to adaptive sampling strategy, where in each case initial samples were randomly chosen. 
Although, in the original work, experimental data and simulation data were collocated at the same operating conditions, for adaptive sampling method this is not a required condition. 

\begin{figure}[t!]
	\begin{center}
		\begin{subfigure}[t]{0.45\textwidth}
			\includegraphics[width=\textwidth]{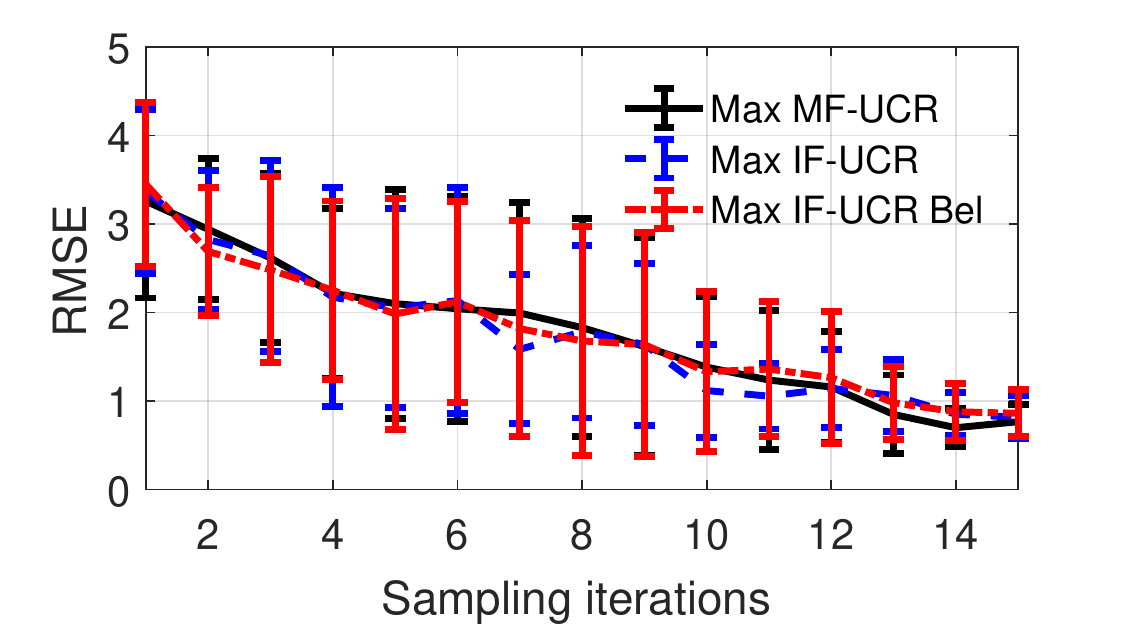}
			\caption{ROOT MEAN SQUARED ERROR (RMSE) AS FUNCTION OF SAMPLING ITERATION}
		\end{subfigure}
		\begin{subfigure}[t]{0.45\textwidth}
			\includegraphics[width=\textwidth]{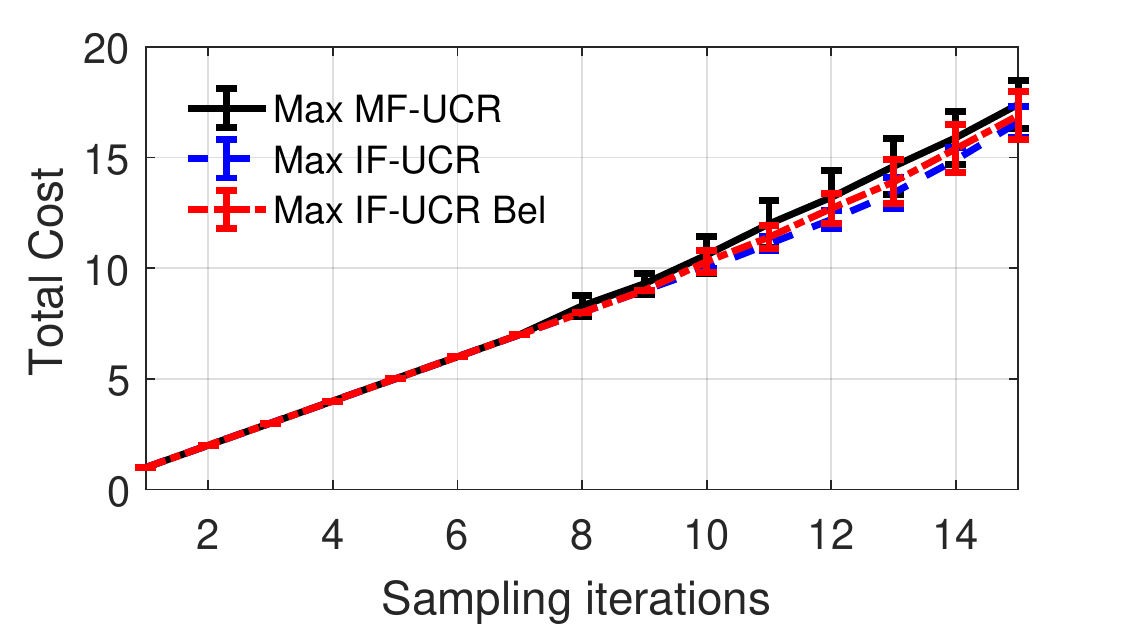}
			\caption{TOTAL COST AS FUNCTION OF SAMPLING ITERATION}
		\end{subfigure}
	\end{center}
	\caption{COMPARISON OF METHODS FOR FLUIDIZED BED PROCESS WITH COST RATIO OF HIGH-FIDELITY AND LOW-FIDELITY OF 2:1}
	\label{fig:FB_1_2}
\end{figure}

\begin{figure}[t!]
	\begin{center}
		\begin{subfigure}[t]{0.45\textwidth}
			\includegraphics[width=\textwidth]{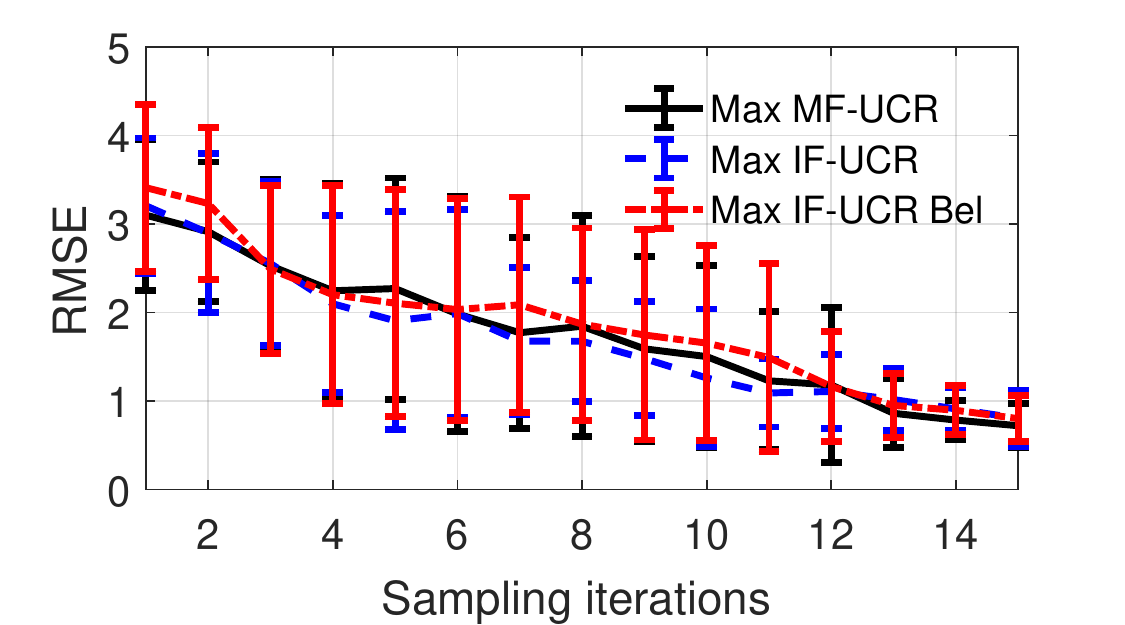}
			\caption{ROOT MEAN SQUARED ERROR (RMSE) AS FUNCTION OF SAMPLING ITERATION}
		\end{subfigure}
		\begin{subfigure}[t]{0.45\textwidth}
			\includegraphics[width=\textwidth]{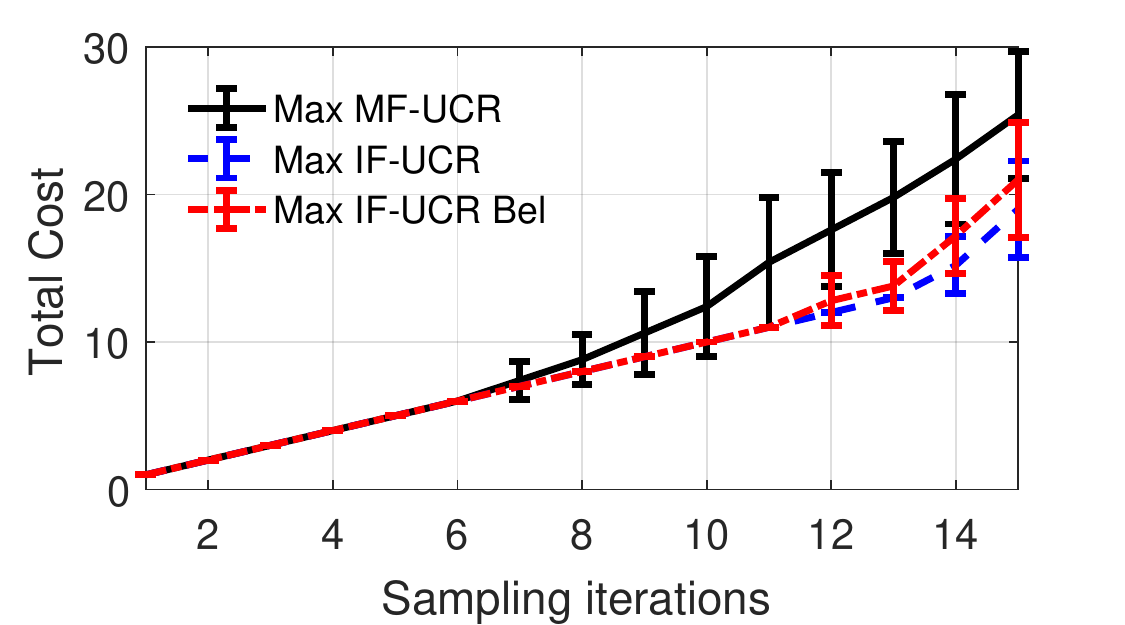}
			\caption{TOTAL COST AS FUNCTION OF SAMPLING ITERATION}
		\end{subfigure}
	\end{center}
	\caption{COMPARISON OF METHODS FOR FLUIDIZED BED PROCESS WITH COST RATIO OF HIGH-FIDELITY AND LOW-FIDELITY OF 5:1}
	\label{fig:FB_1_5}
\end{figure}

\begin{figure}[t!]
	\begin{center}
		\begin{subfigure}[t]{0.45\textwidth}
			\includegraphics[width=\textwidth]{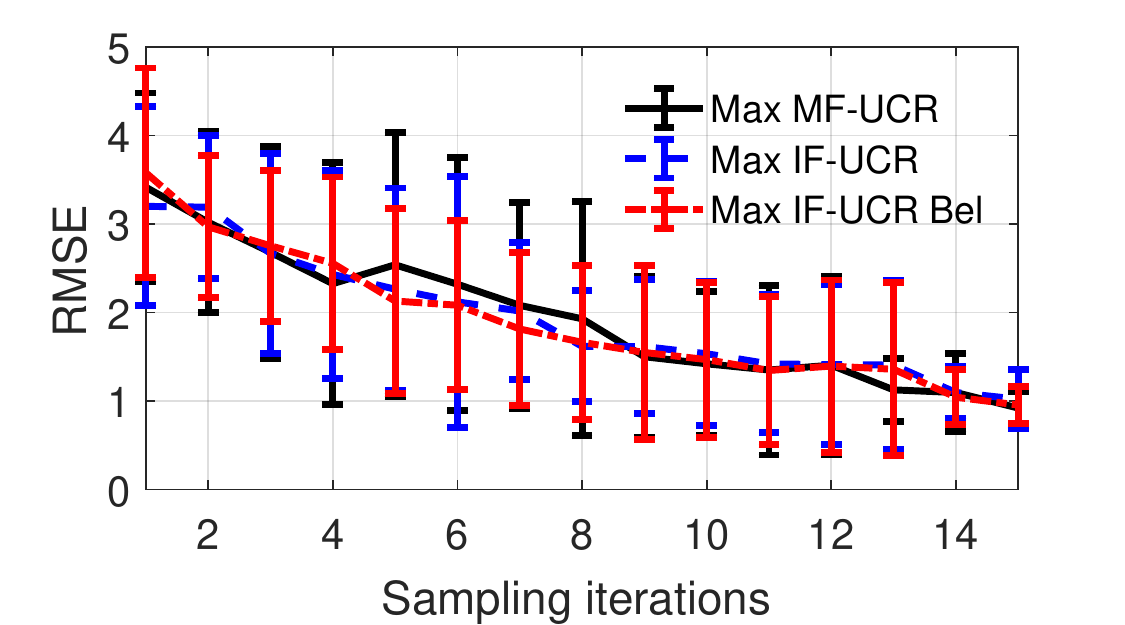}
			\caption{ROOT MEAN SQUARED ERROR (RMSE) AS FUNCTION OF SAMPLING ITERATION}
		\end{subfigure}
		\begin{subfigure}[t]{0.45\textwidth}
			\includegraphics[width=\textwidth]{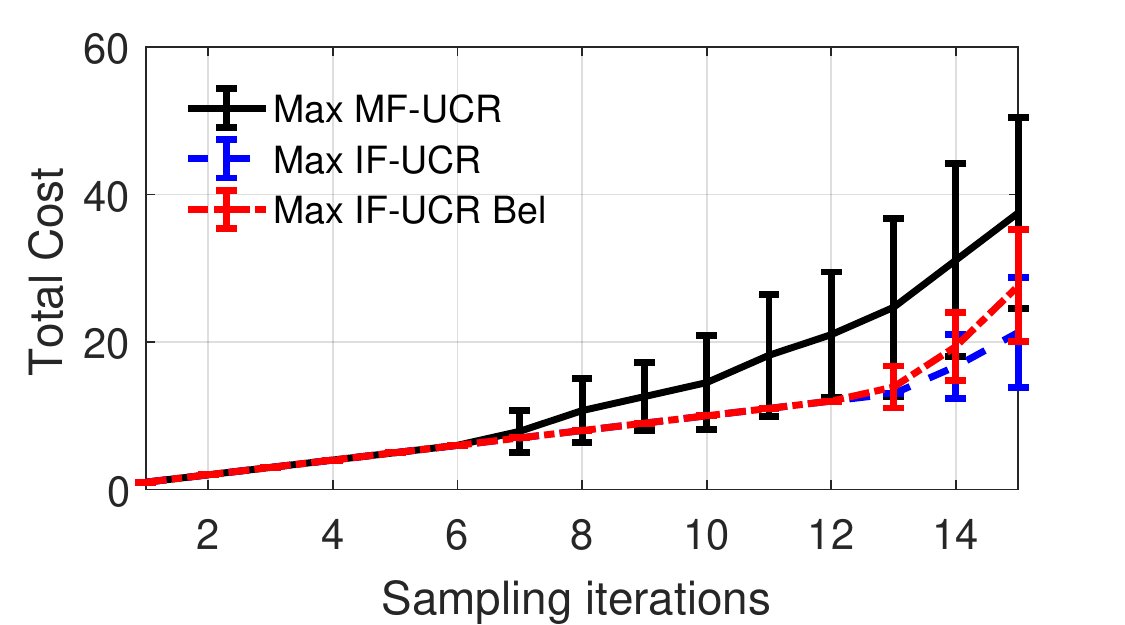}
			\caption{TOTAL COST AS FUNCTION OF SAMPLING ITERATION}
		\end{subfigure}
	\end{center}
	\caption{COMPARISON OF METHODS FOR FLUIDIZED BED PROCESS WITH COST RATIO OF HIGH-FIDELITY AND LOW-FIDELITY OF 10:1}
	\label{fig:FB_1_10}
\end{figure}

Three scenarios are studied with different high-fidelity to low-fidelity cost ratio ($C_H:C_L$) of $2:1$, $5:1$, and $10:1$ and $15$ iterations of adaptive sampling were carried out.
The results are shown in Fig. \ref{fig:FB_1_2}, \ref{fig:FB_1_5}, and \ref{fig:FB_1_10}.
For all the cases, the RMSE converges to similar range for all the adaptive sampling strategy. 
As in the previous tests, difference in total cost has been found to be not very significant for  $C_H:C_L = 2:1$. 
With the increase in cost ratio, the difference in cost has been found to be increasing. 
For the scenario, Max IF-UCR has been found to give the best results in terms of cost, however for higher cost ratio both Max IF-UCR and Max IF-UCR bel has been found to give close result.

\section*{CONCLUSION}
Multi-fidelity Gaussian Process has been commonly used to incorporate cheap low-fidelity data with expensive high-fidelity data to improve the prediction capability of Gaussian process. 
If the low-fidelity analysis is very cheap when compared to high-fidelity analysis, then a straightforward strategy is run large number of low-fidelity analyses to build a very accurate model of low-fidelity analysis ($\eta(x)$) and then adaptively sample only high-fidelity analysis until required accuracy of high fidelity ($y_{mf}(x) = \eta(x) + \delta(x)$) is achieved or cost is within some budget.
On the other hand, if the cost of low-fidelity and high-fidelity analyses are not different, then it is apparent to run only high-fidelity analysis and build only a single-fidelity GP using adaptive sampling strategy. 
In both these cases, adaptive sampling strategy for a single-fidelity GP is sufficient to work.
However, for scenarios when the high-fidelity to low-fidelity cost ratio is not very high or not close to one, then a multi-fidelity adaptive sampling strategy is required to efficiently selecting input space as well as fidelity of the analyses to maximize the information gain with minimum cost. 
In this work a adaptive sampling criteria for selecting the new design point and fidelity of analysis using Maximum Individual Fidelity Uncertainty to Cost Ratio (Max IF-UCR) is demonstrated. 
The criteria is also extended by using Multi-fidelity GP "Beleiver" (Max IF-UCR bel) to carry out adaptively sampling and is compared with baseline case of Max Multi-fidelity  Uncertainty to Cost Ratio (Max MF-UCR).
The method is tested with two analytical test problem and one engineering problem. 
It has been found that when high-fidelity to low-fidelity cost ratio is low ($2:1$), then the proposed approach does not give much cost benefit. 
However, at higher cost ratio ($5:1$ and $10:1$), both Max IF-UCR and Max IF-UCR bel are significantly better in terms of total cost when compared to baseline case of Max MF-UCR.
Although  Max IF-UCR method has been found to be best in all the cases, but Max IF-UCR bel converges to similar trend for higher cost ratios.

\bibliographystyle{unsrt}  

\bibliography{asme2e}


\end{document}